\pdfoutput=1
\documentclass[10pt,twocolumn,letterpaper]{article}

\usepackage{cvpr}
\usepackage{times}
\usepackage{epsfig}
\usepackage{graphicx}
\usepackage{amsmath}
\usepackage{amssymb}
\usepackage{subfloat}
\usepackage{subfig}

\usepackage[breaklinks=true,bookmarks=false,colorlinks]{hyperref}

\cvprfinalcopy % *** Uncomment this line for the final submission

\def\cvprPaperID{****} % *** Enter the CVPR Paper ID here

\begin{document}

%%%%%%%%% TITLE
\title{Coarse-to-fine Face Alignment with Multi-Scale Local Patch Regression}

\author{
Zhiao Huang\\
Megvii Inc.\\
{\tt\small hza@megvii.com}
\and
Erjin Zhou\\
Megvii Inc.\\
{\tt\small zej@megvii.com}
\and
Zhimin Cao\\
Megvii Inc.\\
{\tt\small czm@megvii.com}
}

\maketitle

%%%%%%%%% ABSTRACT
\begin{abstract}
Facial landmark localization plays an important role in face recognition and analysis applications.
In this paper, we give a brief introduction to a coarse-to-fine pipeline with neural networks and sequential regression.
First, a global convolutional network is applied to the holistic facial image to give an initial landmark prediction.
A pyramid of multi-scale local image patches is then cropped to feed to a new network for each landmark to refine the prediction.
As the refinement network outputs a more accurate position estimation than the input, such procedure could be repeated several times until the estimation converges. We evaluate our system on the 300-W dataset~\cite{sagonas2013300} and it outperforms the recent state-of-the-arts.
\end{abstract}

%%%%%%%%% BODY TEXT
\section{Introduction}
The performance of face recognition and analysis applications heavily depends on the effectiveness of facial landmark localization~\cite{taigman2014deepface,sun2014deeply,fan2014learning,zhou2015naive},
which seeks to find the accurate positions for a group of fiducial points pre-defined on the face.
Though under constrained settings, where the face image are without large out-of-plane tilting and occlusion, this task has been considered solved, more general "in-the-wild" cases, where face images are with large pose, illumination, and expression variations, are still regarded as a difficult problem.
Recently, as several challenging facial landmark localization benchmarks proposed, a lot of works have been published and demonstrated promising results under the "in-the-wild" settings~\cite{sun2013deep,zhou2013extensive,burgos2013robust,ren2014face,cao2014face,tzimiropoulos2014gauss,asthana2014incremental,kazemi2014one,zhu2015face,lee2015face,zhang2015learning}.

The coarse-to-fine regression framework has been proposed in the recent approaches~\cite{sun2013deep,zhou2013extensive,zhu2015face}. It tries to estimate the facial landmark positions by a sequence of regression models. In this paper, we present an end-to-end solution under this framework.
Initially, a single neural network is used to predict the facial landmarks holistically.
A subsequent geometric correction is applied to turn the face image to a canonical form according to the estimated scale and rotation.
Then a series of networks are applied to refine each landmark's position estimation.
Each network takes a pyramid of multi-scale local image patches surrounding the landmark as input and outputs a more accurate position estimation. Such refinement is repeated until convergence.
We evaluate our method on the popular 300-W~\cite{sagonas2013300} dataset.
The results show that the proposed approach outperforms the state-of-the-arts by a remarkable margin.

\begin{figure}
\centering
\includegraphics[scale=0.25]{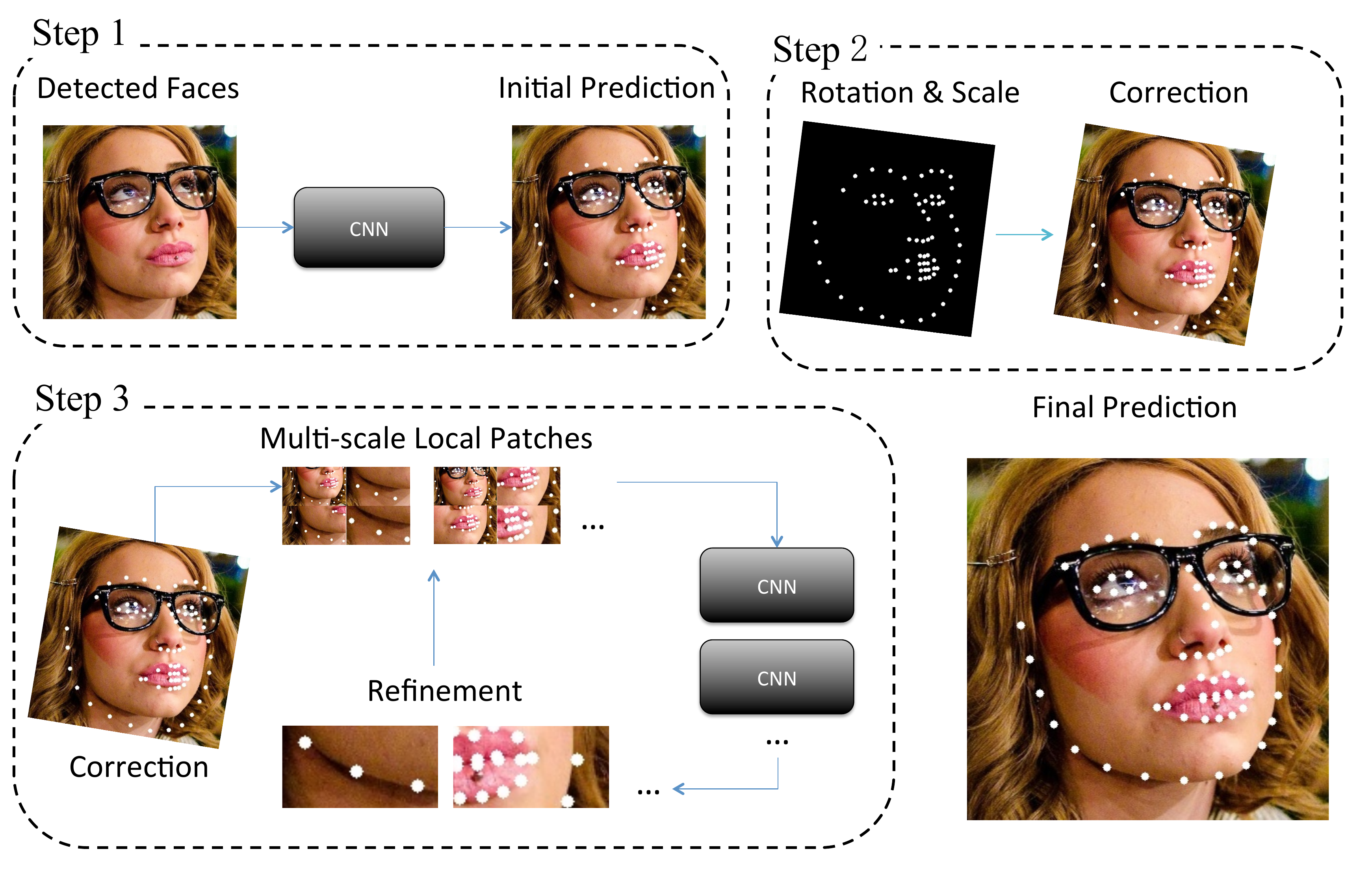}
\caption{
{\bf System overview.} Given a face image, a single neural network predicts an initial estimation.
A subsequent geometric transformation is applied to correct the face image to a canonical form according to the estimated face scale and rotation.
Finally, a series of networks crop multi-scale image patches as inputs and refine the landmarks sequentially until the prediction converges.
}
\label{fig:sys}
\end{figure}
\section{Proposed Method}
In Fig.~\ref{fig:sys}, we illustrate our coarse-to-fine landmark localization pipeline with three steps.

{\bf Step 1.} An off-the-shelf face detector is used to give an initial bounding box of the face. 
The the face region patch is cropped and scaled to a normalized size. 
A global deep convolutional network taking the normalized image patch as input predicts the landmark positions. 
%In the experiments, we find that making the initial bounding box cover larger scope improves the results since it helps the network to capture the global structure information about the face.

{\bf Step 2.} The in-plane rotation and scaling of face are corrected.
After getting a global estimate from the Step 1, the size of the face (more accurate than then face bounding box provided by face detector) and the in-plane face rotation angle is calculated.
Then we canonicalize the face image according to the estimated rotation and scale by applying a similarity transformation.

{\bf Step 3.}
A series of networks refine the landmark estimate sequentially until convergence.
Each network $i$ takes a multi-scale image pyramid as input and attends to refine a set of landmarks $s_i$, which have a spatial or semantic relationships in the face.
Each landmark is predicted by at least one network.
We average the prediction of each landmark obtained from multiple networks to compute the final result.
We sequentially apply this refinement until the prediction converges.

\section{Experiments and Comparisons}
\subsection{Datasets}
{\bf  Megvii Facial Landmark Database.}
We collect a large amount of data from Internet, referred to as Megvii Facial Landmark Database (MFLD).
It contains about 21,000 faces with manually labeled 81 landmark points.
We select 20,000 faces as the training set and leave the remains as test set.

{\bf 300-W Dataset.}
The 300-W dataset~\cite{sagonas2013300} is a popular face alignment benchmark which contains faces collected in-the-wild with large pose, illumination, and expression variants.
The dataset consists 3,148 training images and 689 test images with 68 labeled  landmarks.
The test images are evaluated with three part: common set, challenging set and full set.
The common set contains the 554 images from test set of LFPW and HELEN and the challenging set contains 135 images from iBUG.
The full set is the union of common and challenging sets.

\subsection{Results}
We train our system on the MFLD to output a estimation of 81 predefined landmark coordinates.
Given the landmark definition mismatch between the MFLD and 300-W's dataset, we then apply a linear least-square regression to map the MFLD's 81-coordinates to the 300-W's 68-coordinates.
Finally, we re-train the refinement networks in the Step 3 to further fine-tune the the results.
Both the least-square regression and the refinement networks are trained on the training set of 300-W.

During the evaluation, the performance is measured as the average distance between prediction and ground truth, normalized by the inter-pupil distance.

%We first evaluate the performance of coarse-to-fine strategy in face alignment.
In table~\ref{table:self_res}, we report the mean error of the test set of MFLD through the coarse-to-fine pipeline.
We observe that the refinement steps boost  the performance significantly.
The recursive application of local refinement improves the results further and finally converges.

\begin{figure*}
\centering
\caption{
{\bf Results on the iBUG dataset (the challenging part of 300-W test set).}
Our system predicts accurate results even on the faces with large pose, illumination, and expression variations.
We use face bounding boxes provided by Megvii Face API~\cite{megviiapi}.
The images are sorted from the largest normalized mean error to the smallest.
}
\label{fig:ibug}
\includegraphics[scale=0.17]{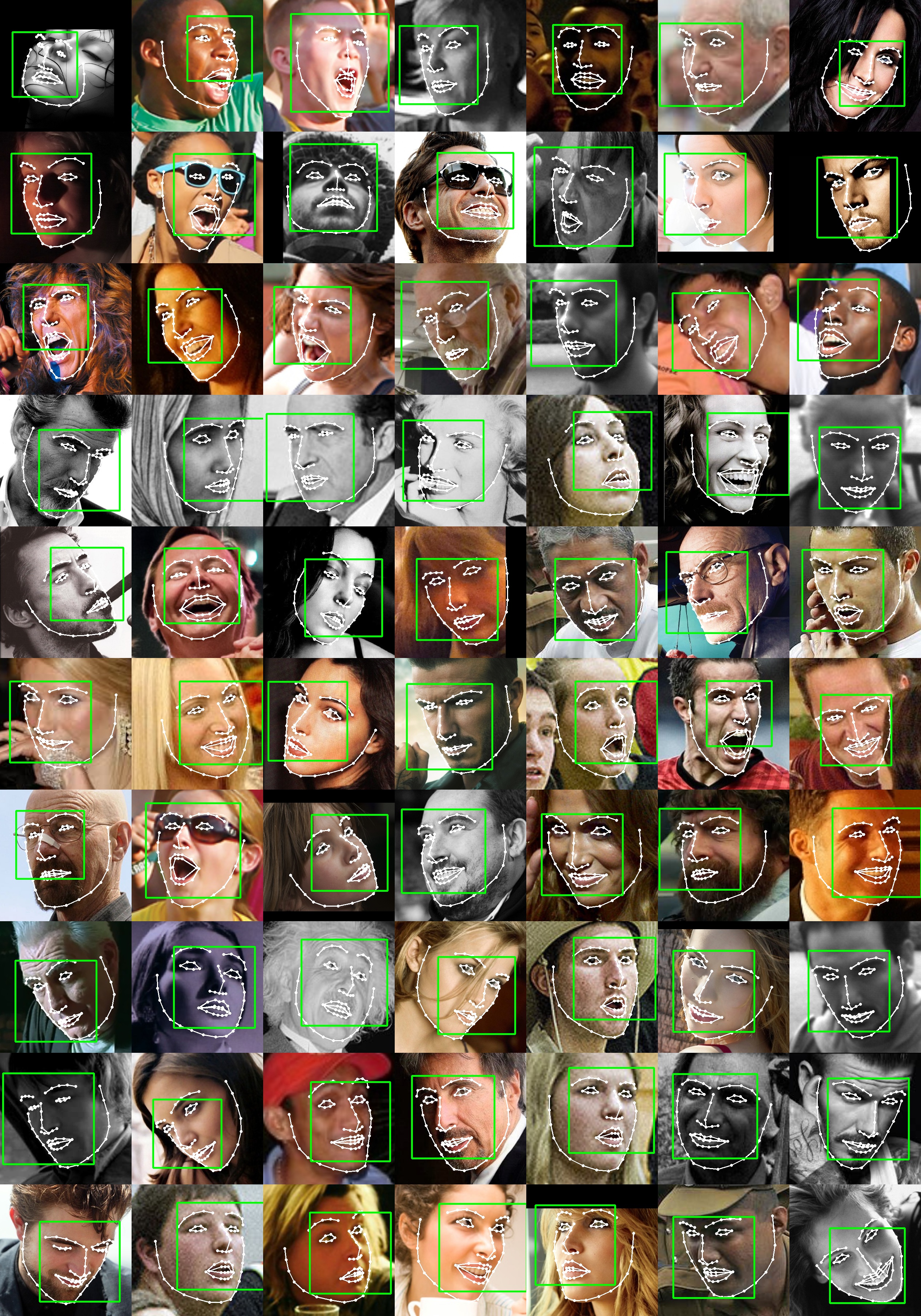}
\end{figure*}

Table~\ref{table:res} shows the comparison results with several recent state-of-the-arts on the 300-W dataset.
We significantly improve the performances and yields the highest accuracy on the 300-W dataset in all settings.
We also present all the prediction results from the challenging part of the 300-W (also named as iBUG dataset) in Fig.~\ref{fig:ibug}.
It can be observed that our system predicts accurate results even with large pose, illumination, and expression on the faces.

\begin{table}[t]
\centering
\caption{
{\bf The normalized mean error on the test set of MFLD through the proposed pipeline.}
}
\label{table:self_res}
\begin{tabular}{|c|c|}\hline
Stage&Mean Error\\\hline
Coarse estimate (Step 1)&5.93\\
Refinement-0 (Step 3)&4.27\\
Refinement-1 (Step 3)&4.16\\
Refinement-2 (Step 3)&4.15\\\hline
\end{tabular}
\end{table}

\begin{table}[t]
\centering
\caption{
{\bf The evaluation on 300-W dataset.}
}
\label{table:res}
\begin{tabular}{|c|c|c|c|}\hline
Methods&Common&Challenging&Fullset\\\hline
Zhu et.al~\cite{zhu2012face}&8.22&18.33&10.20\\
DRMF~\cite{asthana2013robust}&6.65&19.79&9.22\\
ESR~\cite{cao2014face}&5.28&17.00&7.58\\
RCPR~\cite{burgos2013robust}&6.18&17.26&8.35\\
SDM~\cite{xiong2013supervised}&5.57&15.40&7.50\\
Smith et.al~\cite{smith2014nonparametric}&-&13.30&-\\
Zhao et.al~\cite{zhao2014unified}&-&-&6.31\\
GN-DPM~\cite{tzimiropoulos2014gauss}&5.78&-&-\\
CFAN~\cite{zhang2014coarse}&5.50&-&-\\
ERT~\cite{kazemi2014one}&-&-&6.40\\
LBF~\cite{ren2014face}&4.95&11.98&6.32\\
LBF fast~\cite{ren2014face}&5.38&15.50&7.37\\
cGPRT~\cite{lee2015face}&-&-&5.71\\
CFSS~\cite{zhu2015face}&4.73&9.98&5.76\\
CFSS Practical~\cite{zhu2015face}&4.79&10.92&5.99\\
DCR~\cite{lai2015deep}&4.19&8.42&5.02\\
Megvii-Face++&{\bf 3.83}&{\bf 7.46}&{\bf 4.54}\\\hline
\end{tabular}
\end{table}

\section{Conclusions}
In this paper, we propose an end-to-end framework, which predicts the facial landmarks through a coarse-to-fine pipeline with multi-scale local patch regression.
Our results outperform the recent state-of-the-arts on the 300-W dataset.

\begin{figure*}
\centering
\includegraphics[scale=0.17]{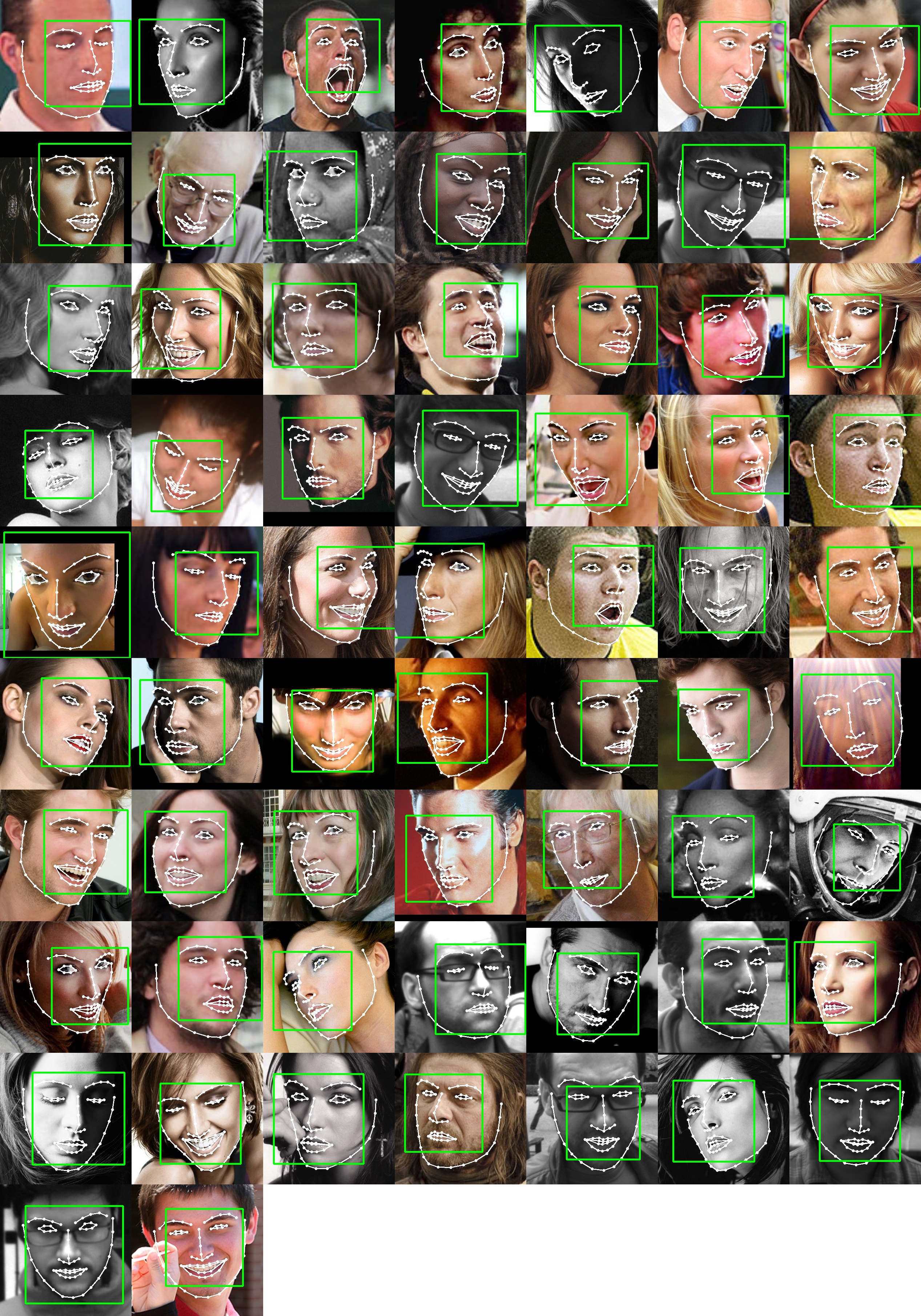}
\newpage
\end{figure*}

{
\bibliographystyle{ieee}
\bibliography{face-alignment-tech-report}

\begin{thebibliography}{10}\itemsep=-1pt

\bibitem{megviiapi}
Megvii face api.
\newblock \url{http://www.faceplusplus.com/}.

\bibitem{asthana2013robust}
A.~Asthana, S.~Zafeiriou, S.~Cheng, and M.~Pantic.
\newblock Robust discriminative response map fitting with constrained local
  models.
\newblock In {\em Computer Vision and Pattern Recognition (CVPR), 2013 IEEE
  Conference on}, pages 3444--3451. IEEE, 2013.

\bibitem{asthana2014incremental}
A.~Asthana, S.~Zafeiriou, S.~Cheng, and M.~Pantic.
\newblock Incremental face alignment in the wild.
\newblock In {\em Computer Vision and Pattern Recognition (CVPR), 2014 IEEE
  Conference on}, pages 1859--1866. IEEE, 2014.

\bibitem{burgos2013robust}
X.~P. Burgos-Artizzu, P.~Perona, and P.~Doll{\'a}r.
\newblock Robust face landmark estimation under occlusion.
\newblock In {\em Computer Vision (ICCV), 2013 IEEE International Conference
  on}, pages 1513--1520. IEEE, 2013.

\bibitem{cao2014face}
X.~Cao, Y.~Wei, F.~Wen, and J.~Sun.
\newblock Face alignment by explicit shape regression.
\newblock {\em International Journal of Computer Vision}, 107(2):177--190,
  2014.

\bibitem{fan2014learning}
H.~Fan, M.~Yang, Z.~Cao, Y.~Jiang, and Q.~Yin.
\newblock Learning compact face representation: Packing a face into an int32.
\newblock In {\em Proceedings of the ACM International Conference on
  Multimedia}, pages 933--936. ACM, 2014.

\bibitem{kazemi2014one}
V.~Kazemi and J.~Sullivan.
\newblock One millisecond face alignment with an ensemble of regression trees.
\newblock In {\em Computer Vision and Pattern Recognition (CVPR), 2014 IEEE
  Conference on}, pages 1867--1874. IEEE, 2014.

\bibitem{lai2015deep}
H.~Lai, S.~Xiao, Z.~Cui, Y.~Pan, C.~Xu, and S.~Yan.
\newblock Deep cascaded regression for face alignment.
\newblock {\em arXiv preprint arXiv:1510.09083}, 2015.

\bibitem{lee2015face}
D.~Lee, H.~Park, and C.~D. Yoo.
\newblock Face alignment using cascade gaussian process regression trees.
\newblock In {\em Proceedings of the IEEE Conference on Computer Vision and
  Pattern Recognition}, pages 4204--4212, 2015.

\bibitem{ren2014face}
S.~Ren, X.~Cao, Y.~Wei, and J.~Sun.
\newblock Face alignment at 3000 fps via regressing local binary features.
\newblock In {\em Computer Vision and Pattern Recognition (CVPR), 2014 IEEE
  Conference on}, pages 1685--1692. IEEE, 2014.

\bibitem{sagonas2013300}
C.~Sagonas, G.~Tzimiropoulos, S.~Zafeiriou, and M.~Pantic.
\newblock 300 faces in-the-wild challenge: The first facial landmark
  localization challenge.
\newblock In {\em Computer Vision Workshops (ICCVW), 2013 IEEE International
  Conference on}, pages 397--403. IEEE, 2013.

\bibitem{smith2014nonparametric}
B.~M. Smith, J.~Brandt, Z.~Lin, and L.~Zhang.
\newblock Nonparametric context modeling of local appearance for pose-and
  expression-robust facial landmark localization.
\newblock In {\em Computer Vision and Pattern Recognition (CVPR), 2014 IEEE
  Conference on}, pages 1741--1748. IEEE, 2014.

\bibitem{sun2013deep}
Y.~Sun, X.~Wang, and X.~Tang.
\newblock Deep convolutional network cascade for facial point detection.
\newblock In {\em Computer Vision and Pattern Recognition (CVPR), 2013 IEEE
  Conference on}, pages 3476--3483. IEEE, 2013.

\bibitem{sun2014deeply}
Y.~Sun, X.~Wang, and X.~Tang.
\newblock Deeply learned face representations are sparse, selective, and
  robust.
\newblock {\em arXiv preprint arXiv:1412.1265}, 2014.

\bibitem{taigman2014deepface}
Y.~Taigman, M.~Yang, M.~Ranzato, and L.~Wolf.
\newblock Deepface: Closing the gap to human-level performance in face
  verification.
\newblock In {\em Computer Vision and Pattern Recognition (CVPR), 2014 IEEE
  Conference on}, pages 1701--1708. IEEE, 2014.

\bibitem{tzimiropoulos2014gauss}
G.~Tzimiropoulos and M.~Pantic.
\newblock Gauss-newton deformable part models for face alignment in-the-wild.
\newblock In {\em Computer Vision and Pattern Recognition (CVPR), 2014 IEEE
  Conference on}, pages 1851--1858. IEEE, 2014.

\bibitem{xiong2013supervised}
X.~Xiong and F.~De~la Torre.
\newblock Supervised descent method and its applications to face alignment.
\newblock In {\em Computer Vision and Pattern Recognition (CVPR), 2013 IEEE
  Conference on}, pages 532--539. IEEE, 2013.

\bibitem{zhang2014coarse}
J.~Zhang, S.~Shan, M.~Kan, and X.~Chen.
\newblock Coarse-to-fine auto-encoder networks (cfan) for real-time face
  alignment.
\newblock In {\em Computer Vision--ECCV 2014}, pages 1--16. Springer, 2014.

\bibitem{zhang2015learning}
Z.~Zhang, P.~Luo, C.~C. Loy, and X.~Tang.
\newblock Learning deep representation for face alignment with auxiliary
  attributes.
\newblock 2015.

\bibitem{zhao2014unified}
X.~Zhao, T.-K. Kim, and W.~Luo.
\newblock Unified face analysis by iterative multi-output random forests.
\newblock In {\em Computer Vision and Pattern Recognition (CVPR), 2014 IEEE
  Conference on}, pages 1765--1772. IEEE, 2014.

\bibitem{zhou2015naive}
E.~Zhou, Z.~Cao, and Q.~Yin.
\newblock Naive-deep face recognition: Touching the limit of lfw benchmark or
  not?
\newblock {\em arXiv preprint arXiv:1501.04690}, 2015.

\bibitem{zhou2013extensive}
E.~Zhou, H.~Fan, Z.~Cao, Y.~Jiang, and Q.~Yin.
\newblock Extensive facial landmark localization with coarse-to-fine
  convolutional network cascade.
\newblock In {\em Computer Vision Workshops (ICCVW), 2013 IEEE International
  Conference on}, pages 386--391. IEEE, 2013.

\bibitem{zhu2015face}
S.~Zhu, C.~Li, C.~C. Loy, and X.~Tang.
\newblock Face alignment by coarse-to-fine shape searching.
\newblock In {\em Proceedings of the IEEE Conference on Computer Vision and
  Pattern Recognition}, pages 4998--5006, 2015.

\bibitem{zhu2012face}
X.~Zhu and D.~Ramanan.
\newblock Face detection, pose estimation, and landmark localization in the
  wild.
\newblock In {\em Computer Vision and Pattern Recognition (CVPR), 2012 IEEE
  Conference on}, pages 2879--2886. IEEE, 2012.

\end{thebibliography}
}

\end{document}